\documentclass[11pt]{article}
\pdfoutput=1
\usepackage{acl2015}
\usepackage{times}
\usepackage{latexsym}
\usepackage{amsmath}
\usepackage{multirow}
\usepackage{url}
\usepackage{graphicx}
\usepackage{enumitem}

\usepackage{pdfsync}
\usepackage{graphicx}
\usepackage{booktabs}
\usepackage{amssymb}
\usepackage{multicol}
\usepackage{watermark}
\usepackage{bm}
\usepackage{xspace}
\usepackage[export]{adjustbox}
\usepackage[utf8]{inputenc}
\usepackage[greek,english]{babel}
\usepackage{CJKutf8}

\newcommand{\toggle}[1]{}
\renewcommand{\toggle}[1]{#1} 

\newif\ifcomment\commenttrue
\ifcomment
\newcommand{\macomment}[1]{\marginpar{\begin{center}\textcolor{orange}{#1}\end{center}}}

\else
\newcommand{\macomment}[1]{} 
\fi

\def\textit#1{{\it #1}}
\def\textbf#1{{\bf #1}}
\def\textsl#1{{\sl #1}}
\def\texttt#1{{\tt #1}}

\renewcommand{\vec}[1]{\bm{#1}}
\newcommand\vs{\vec{s}}

\newcommand\va{\vec{a}}
\newcommand\vb{\vec{b}}
\newcommand\vc{\vec{c}}
\newcommand\vd{\vec{d}}

\newcommand\vg{\vec{g}}
\newcommand\vh{\vec{h}}

\newcommand\vq{\vec{q}}

\newcommand\vv{\vec{v}}
\newcommand\vw{\vec{w}}
\newcommand\vsd{\vec{sd}}
\newcommand\vy{\vec{y}}
\newcommand\vx{\vec{x}}
\newcommand\vr{\vec{r}}
\newcommand\vz{\vec{z}}
\newcommand\vW{\vec{W}}
\newcommand\vU{\vec{U}}
\newcommand\vV{\vec{V}}

\newcommand\vQ{\vec{Q}}

\newcommand{\unk}{$\langle$UNK$\rangle$\xspace}

\title{Implicit Distortion and Fertility Models for \\ Attention-based Encoder-Decoder NMT Model}

\toggle{
\author{Shi Feng$^{\dag}$ \\
  Shanghai Jiao Tong University\\
  Shanghai, P.R. China\\
  {\tt sjtufs@gmail.com} \\\And
  Shujie Liu, Mu Li, Ming Zhou\\
  Microsoft Research\\
  Beijing, P.R. China\\
  {\tt shujliu,muli,mingzhou@microsoft.com}
  }
}

\date{}

\begin{document}
\maketitle

{\let\thefootnote\relax\footnotetext{$^\dag$Work done while Shi was an intern at Microsoft Research.}}

\begin{abstract}
Neural machine translation has shown very promising results lately.
Most NMT models follow the encoder-decoder framework.
To make encoder-decoder models more flexible, attention mechanism was introduced to machine translation and also other tasks like speech recognition and image captioning.
We observe that the quality of translation by attention-based encoder-decoder can be significantly damaged when the alignment is incorrect. We attribute these problems to the lack of distortion and fertility models.
Aiming to resolve these problems, we propose new variations of attention-based encoder-decoder and compare them with other models on machine translation. Our proposed method achieved an improvement of 2 BLEU points over the original attention-based encoder-decoder.
\end{abstract}

\section{Introduction}
\label{section:intro}

Neural machine translation has shown promising results lately. Most NMT methods follow the encoder-decoder framework proposed by \cite{cho2014learning}, which typically consists of two RNNs: the encoder RNN reads the source sentence and transform it into vector representation; the decoder RNN takes the vector representation and generates the target sentence word by word. The decoder will stop once a special symbol denoting the end of the sentence is generated. This encoder-decoder framework can be used on general sequence-to-sequence tasks \cite{sutskever2014sequence}, like question answering and text summarization. After some modification, for example replacing the RNN encoder with a CNN, the model can also be applied to tasks like image captioning \cite{vinyals2014show,xu2015show}. In the following discussion, we focus on the task of machine translation.

In the original encoder-decoder model, although the encoder RNN generates a set of hidden states, one at each position of the source sentence, the decoder only takes the last one. This design in effect compresses the variable-length source sentence into a fixed-length context vector, with the information of each source word implicitly stored in the context vector. Thus the decoder cannot easily make full use of the whole sequence of encoder hidden states. To make it more flexible and generalize the fixed-length representation to a variable-length one, it was proposed to use attention mechanism for machine translation \cite{bahdanau2014neural}.

\begin{figure}[t]
\centering
\includegraphics[width=0.8\columnwidth]{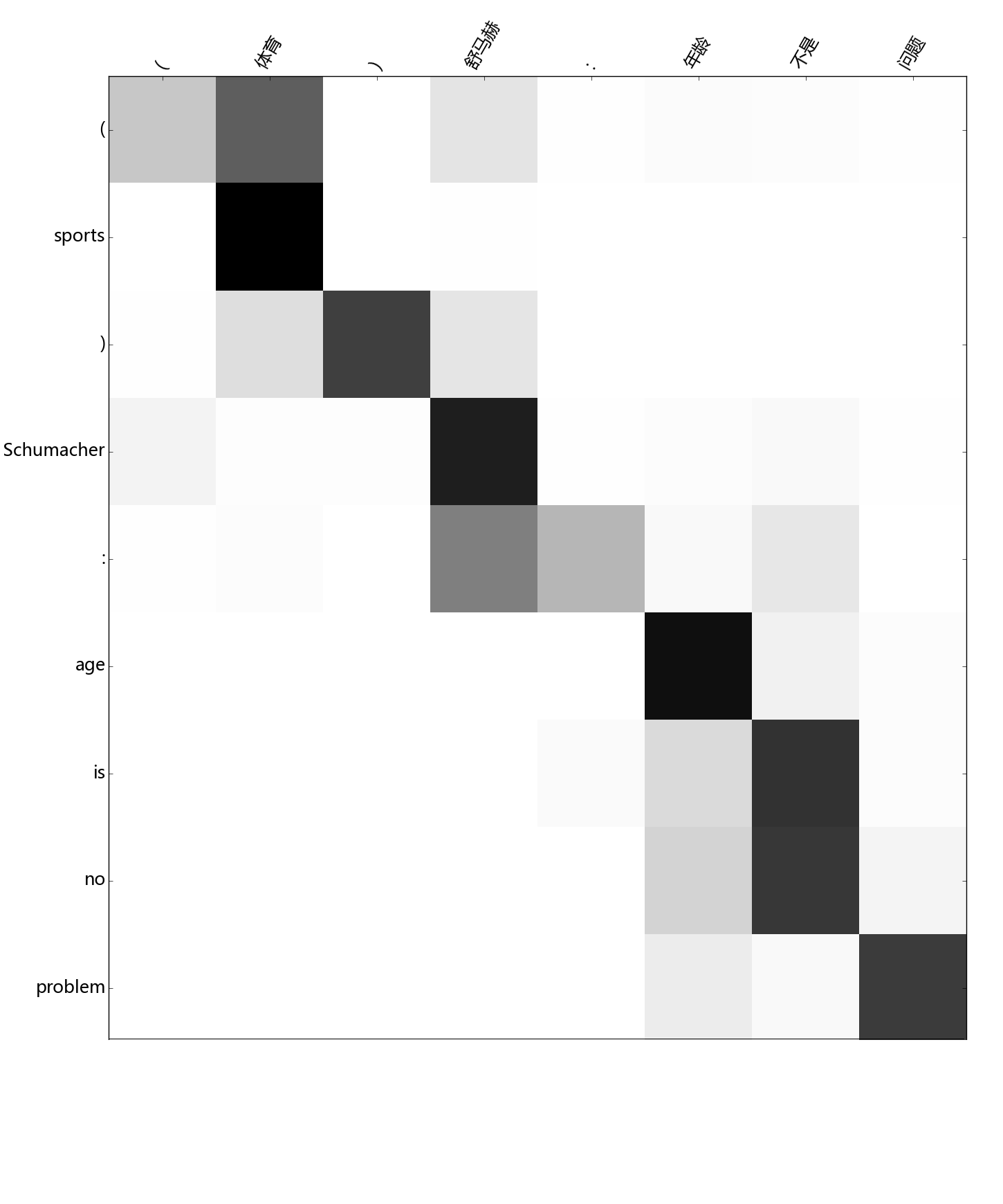}
\caption{Alignment by attention mechanism.
    Each row is a distribution of how the corresponding target word (English) is aligned to source word.
    Darker color denotes higher weight.
}
\label{fig:intro_align_1}
\end{figure}

Attention mechanism was first proposed to allow models to learn alignments between different modalities, e.g., between image objects and agent actions in the dynamic control problem \cite{mnih2014recurrent}. 

In \cite{bahdanau2014neural}, attention mechanism was applied to machine translation to learn an alignment between source words and target words. Fig.~\ref{fig:intro_align_1} shows a sample alignment given by attention mechanism. 

With the ability of learning alignments between different modalities from attention mechanism, attention-based encoder-decoder model is more powerful than just encoder-decoder and has been used for many tasks like question answering \cite{hermann2015teaching}, speech recognition \cite{bahdanau2015end,chorowski2014end}, image captioning \cite{xu2015show} and visual question answering \cite{xu2015ask,chen2015abc,shih2015look}. In these applications, variations of attention mechanism were proposed to enhance its performance. 


\section{Problems of Attention Mechanism}
\label{section:problems}

By training the encoder-decoder model with attention mechanism, we get an alignment from target word to source word. This alignment helps translation by allowing re-ordering. But since the alignment by attention is not always accurate, we observed that in many cases where the alignment is incorrect, the translation quality is significantly damaged. We attribute this kind of problem to the lack of explicit distortion and fertility models in attention-based encoder-decoder model.

\begin{figure}[t]
\centering
\includegraphics[width=1.0\columnwidth]{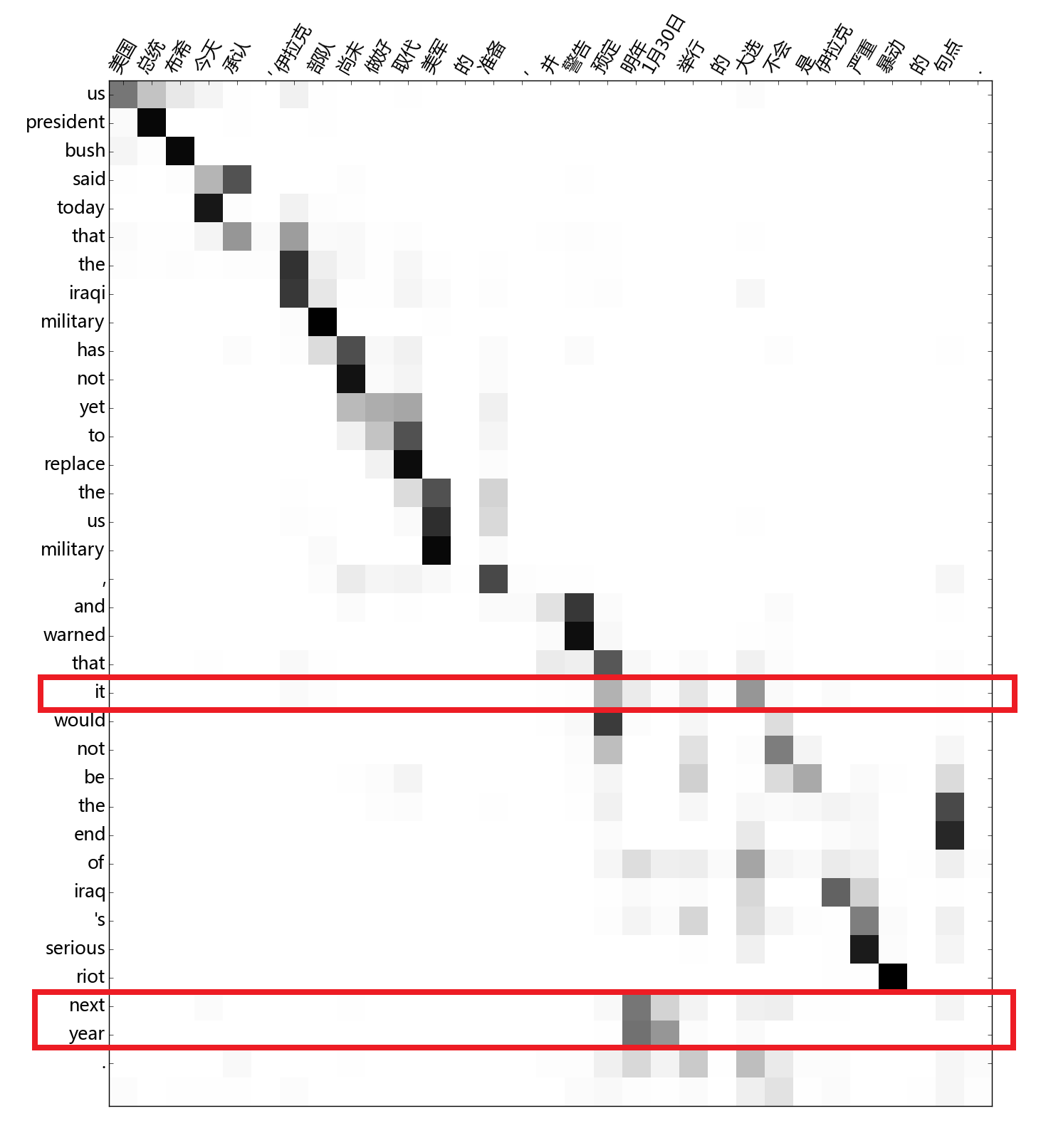}
\caption{Wrong alignment due to lack of distortion model.}
\label{fig:prob_distortion_1}
\end{figure}

\subsection{Lack of Distortion Model}
\label{section:problem_distortion}

In SMT, the distortion model controls how the words are re-ordered. In Fig.~\ref{fig:prob_distortion_1} we show an example alignment given by attention mechanism where incorrect alignment in the middle of the sentence caused the translation to go wrong afterwards. We focus on the later part of the sentence where the correct translation should be:

``...\emph{and warned that the election to be held on january 30th next year would not be and end to serious violence in iraq.}"

which get translated into:

``...\emph{and warned that it would not be the end of iraq's serious violence \textbf{next year}.}"

From the alignment matrix we can see that, the word ``\begin{CJK}{UTF8}{gbsn}预定\end{CJK}" (means ``scheduled") in source sentence is attended to by ``\emph{would}", but the next attention jumped to ``\begin{CJK}{UTF8}{gbsn}大选\end{CJK} (means ``election"), while it should focus on ``\begin{CJK}{UTF8}{gbsn}明年\end{CJK}" (means ``next year") or on the date. After this incorrect re-ordering, the meaning of the source sentence is twisted in the translation. If the attention is aware of the previous alignment, it should be able to order ``\emph{next year}" after ``\emph{election}" and reserve the mearning of source sentence correctly.

In this kind of casese, the translation can go wrong due to incorrect re-ordering. We attribute this kind of problem due to the lack of distortion model in attention-based encoder-decoder.

\subsection{Lack of Fertility Model}
\label{section:problem_fertility}

In SMT, the fertility model controls how many target words are translated from a source word.
We observe two phenomena related to the lack of fertility model in attention-based encoder-decoder: the problem of repetition and the problem of coverage.

\begin{figure}[t]
\centering
\includegraphics[width=1.0\columnwidth]{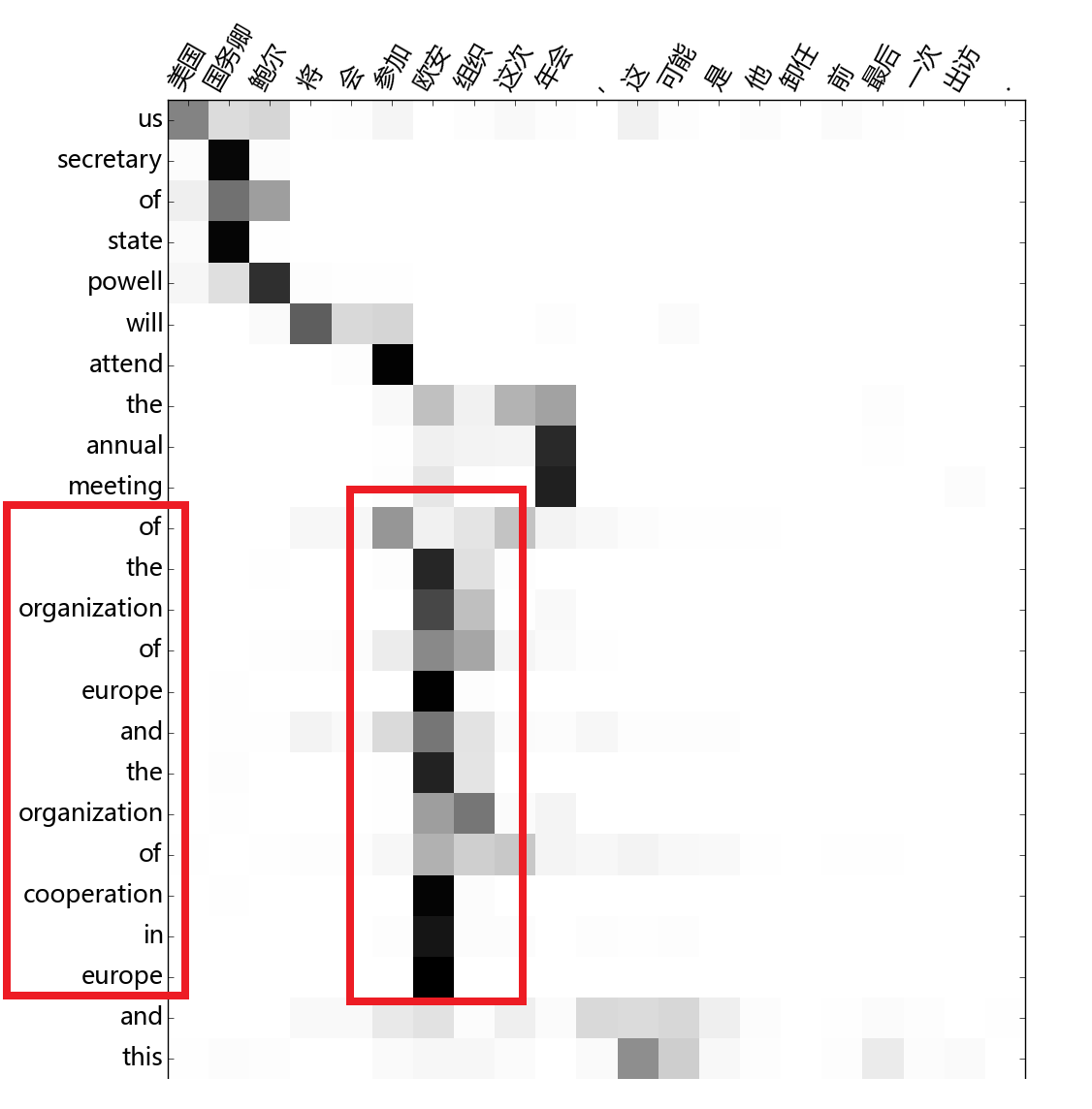}
\caption{Problem of repetition in alignment.}
\label{fig:prob_repetition_1}
\end{figure}

\begin{figure}[t]
\centering
\includegraphics[width=1.0\columnwidth]{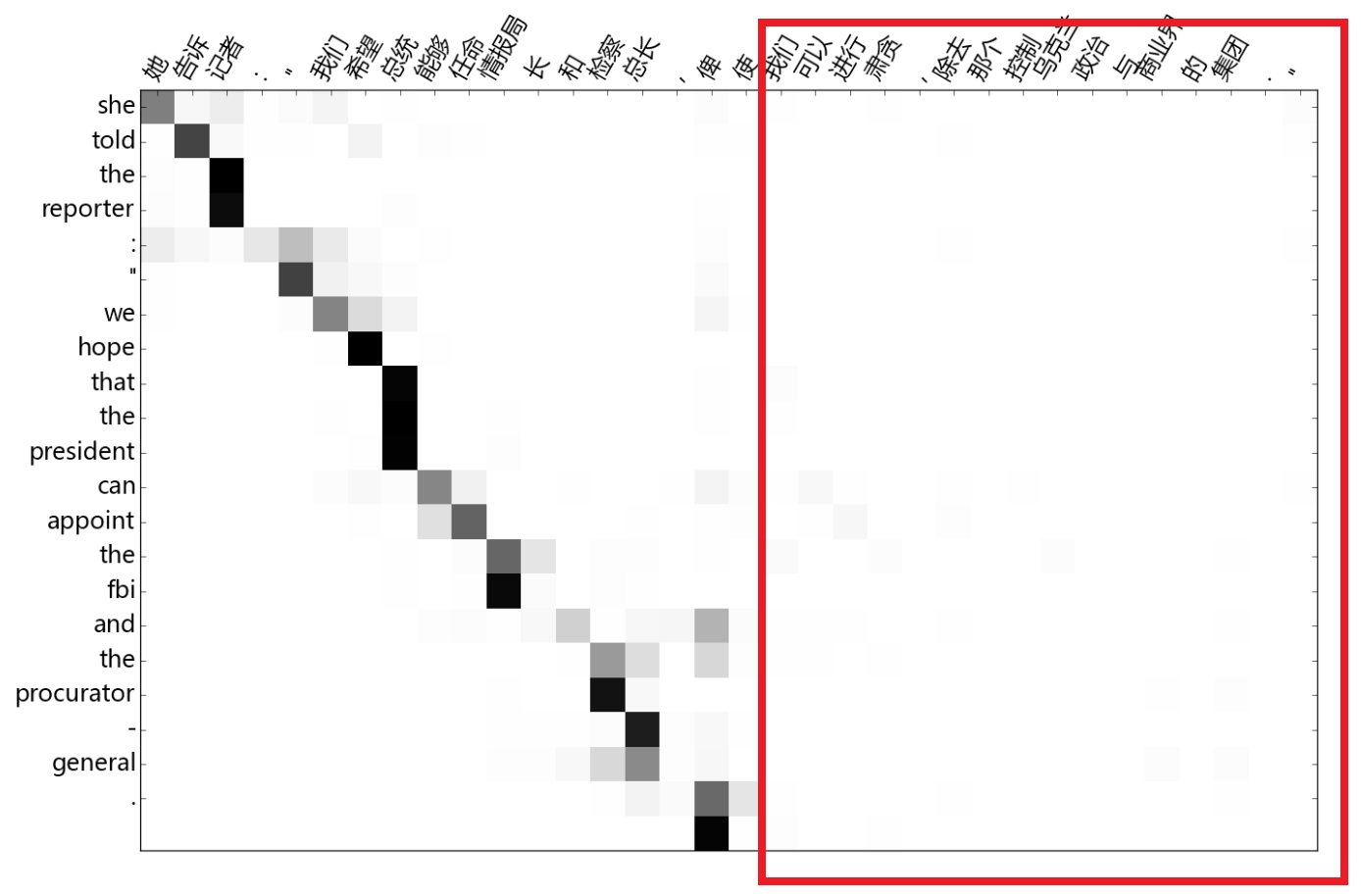}
\caption{Problem of coverage in alignment.}
\label{fig:prob_coverage_1}
\end{figure}

\paragraph{Problem of Repetition}
Fig.~\ref{fig:prob_repetition_1} shows an example of repetition problem in alignment. For consecutive words, the attention mechanism focused on the same position in the source sentence, resulting in repetition in the translation, ``\emph{the organization of europe and the organization of cooperation in europe}".

\paragraph{Problem of Coverage}
Fig.~\ref{fig:prob_coverage_1} shows an example of coverage problem in alignment. We see that some part of the source sentence was not attended to, resulting in significant loss of content in the translation. 

These two problems are due to the lack of fertility model in NMT: in the first case, some source words are translated into too many target words, while in the second case, some sources words are translted into too few target words.

Although attention mechanism already makes the encoder-decoder more flexible by allowing re-ordering, the observed problems demonstrated some restrictions of it. Motivated by these observations, we propose additions of implicit distortion and fertility models to attention-based encoder decoder. In Sec.~\ref{section:recurrent_attention}, we introduce \textsc{RecAtt} and \textsc{RNNAtt} which are designed as an implicit distortion models. In Sec.~\ref{section:conditioned_decoder}, we introduce \textsc{CondDec} which is designed as an implicit fertility model. We verify that the proposed methods can resolve the observed problems in our experiments in Sec.~\ref{section:qualitative}.

\section{Attention-based Encoder-Decoder}
\label{section:methods}

We start by reviewing the RNN used in NMT papers and the encoder-decoder with attention mechanism from \cite{bahdanau2014neural}.

\subsection{Gated Recurrent Unit}
\label{section:gru}

Gated Recurrent Unit (GRU) \cite{cho2014learning} is an RNN alternative similar to LSTM \cite{hochreiter1997long}. It was used in NMT papers \cite{cho2014learning,bahdanau2014neural} and we will use GRU as RNN in our paper. Like normal RNN, GRU computes its hidden state $\vh_i$ based on the input $\vx_i$ and previous hidden state $\vh_{i-1}$:
\begin{align*}
\vh_i = \text{RNN}(\vh_{i-1}, \vx_{i})
\end{align*}

which is computed with update gate and reset gate, formally defined by:
\begin{align*}
  \vr_i &= \sigma(\vW^r\vx_{i} + \vU^r\vh_{i-1}) \\
  \vh'_i &= \tanh(\vr_i \circ \vU\vh_{i-1} + \vW\vx_{i}) \\
  \vz_i &= \sigma(\vW^z\vx_{i} + \vU^z\vh_{i-1}) \\
  \vh_i &= (1 - \vz_i) \circ \vh'_i + \vz_i \circ \vh_{i-1}
\end{align*}

where $\vx_{i}$ is the input, $\vh_{i-1}$ is the previous hidden state. $\vz_i$ and $\vr_i$ are the values of update gate and reset gate respectively. $\circ$ denotes bit-wise product.
Biases are dropped for simplicity.

\subsection{RNNSearch \cite{bahdanau2014neural}}
\label{section:rnnsearch}

\paragraph{Encoder}

The encoder used in \textsc{RNNSearch} \cite{bahdanau2014neural} is a bi-directional RNN. It consists of two independent RNNs, one reading the source sentence from left to right, another from right to left:
\begin{align*}
  \overrightarrow{\vs}_i &= \text{RNN}(\overrightarrow{\vs}_{i-1}, \vx_i) \\
  \overleftarrow{\vs}_i &= \text{RNN}(\overleftarrow{\vs}_{i+1}, \vx_i) 
\end{align*}

where $\vx_i$ is the word embedding of source word at position $i$. The representation at position $i$ is then defined as the concatenation of $\overrightarrow{\vs}_i$ and $\overleftarrow{\vs}_i$: 
\begin{align*}
\vs_i=\left[ \begin{array}{c} \overrightarrow{\vs}_i \\ \overleftarrow{\vs}_i \end{array} \right]
\end{align*}

\paragraph{Decoder with Attention}

Unlike the decoder from \cite{cho2014learning} which takes only the last representation, the decoder with attention mechanism can make full use of the whole representation set ${\vs_j}$. The decoder is assisted by a unit that provides a dynamic context $\vc_i$:
\begin{align*}
  \vc_i &= \text{ATT}(\vh_{i-1}, \{\vs_j\}) \\
\end{align*}

At each decoder step, the attention unit takes both the previous decoder hidden state $\vh_{i-1}$ and the set of encoder representations $\{\vs_j\}$ as input, outputs a weighted average of encoder hidden states as the context $\vc_i$. It uses a match function $\alpha$ to match $\vh_{i-1}$ with each $\vs_j$ and generates the weight $w_{ij}$ for $\vs_j$.
\begin{align*}
  e_{ij} &= \vv^T\tanh\alpha(\vh_{i-1}, \vs_j) \\
  w_{ij} &= \frac{\text{exp}(e_{ij})}{\sum_{k}{\text{exp}(e_{ik})}} \\
  \vc_i &= \sum_{j}{w_{ij}\vs_j}
\end{align*}

The match function can take on many forms, which is analyzed in \cite{luong2015effective}.


In our paper we use the sum match function, which is a more common choice as used in \cite{bahdanau2014neural}.
\begin{align*}
  \alpha_{\text{sum}}(\va, \vb) = \vW^{\alpha}\va + \vU^{\alpha}\vb
\end{align*}

The context $\vc_i$ is then used by the decoder:
\begin{align*}
  \vh_i &= \text{RNN}(\vh_{i-1}, \vy_{i-1}, \vc_i)
\end{align*}

where $\vy_{i-1}$ is the embedding of previous target word.

To predict a target word at position $i$, the decoder state $\vh_i$ is concatenated with $\vc_i$ and $\vy_{i-1}$ and fed through deep out \cite{pascanu2012difficulty} with a single maxout hidden layer \cite{goodfellow2013maxout}, followed by a softmax. We follow this structure in this paper.

\section{Recurrent Attention Mechanisms}
\label{section:recurrent_attention}

\begin{figure}[t]
\centering
\includegraphics[width=0.9\columnwidth]{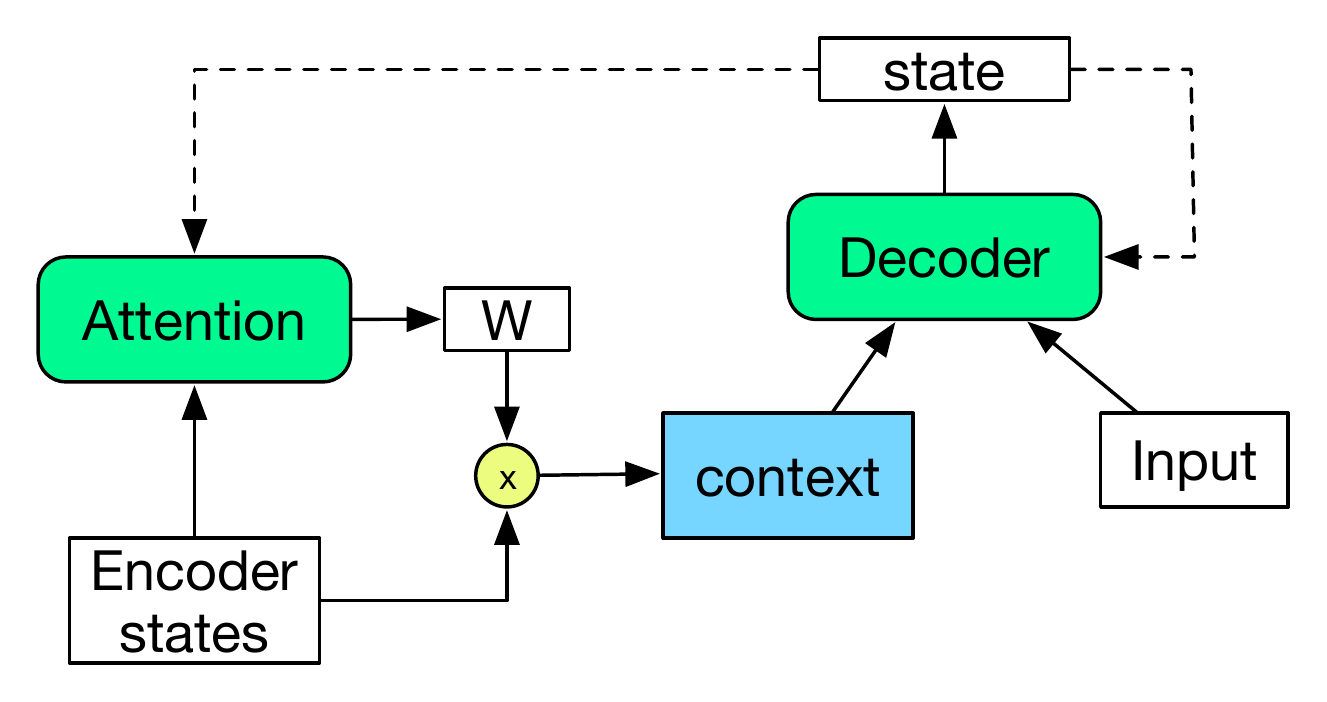}
\caption{Attention Decoder.
    The dashed lines show how the current hidden state is passed to the next decoding step.
    The current attention is computed with the previous hidden state.
}
\label{fig:attention_decoder}
\end{figure}

In Fig.~\ref{fig:attention_decoder} we show an abstraction of the decoder-attention structure.

We note that attention mechanism treats the encoder states as a set, not a sequence, while the source sentence order is crucial to re-ordering. And the state of re-ordering given to the attention unit is all embedded in the hidden state of the decoder - the attention unit itself does not have memory. 

Motivated by the analysis in Sec.~\ref{section:problem_distortion}, we propose to add recurrent paths to the decoder-attention structure to provide the attention unit with more information the re-ordering. With these recurrent paths, instead of making the decoder remembering what the state of re-ordering is, recurrent attention mechanism explicitly keeps track of this information. 

\subsection{RecAtt}
\label{section:recatt}

\begin{figure}[t]
\centering
\includegraphics[width=0.9\columnwidth]{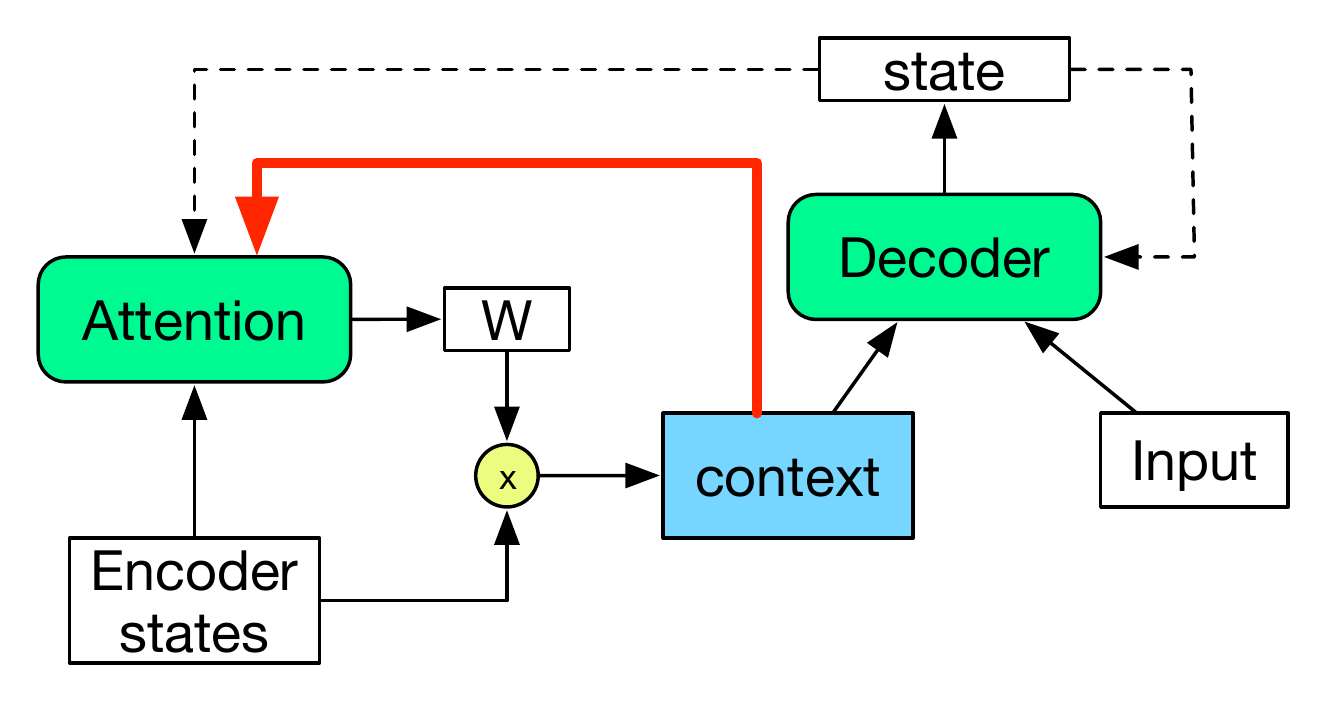}
\caption{RecAtt Decoder.
    The red thick line denotes the recurrent attention path which passes previous attention
    generated context to the attention unit.
}
\label{fig:recatt}
\end{figure}

In this section we introduce our proposed recurrent attention mechanism, RecAtt.

We pass the previous context directly to the attention unit to inform it about the alignment from the previous step. 

The decoder with RecAtt follows:
\begin{align*}
  \vc_i &= \text{ATT}(\vh_{i-1}, \vc_{i-1} \{\vs_j\}) \\
  \vh_i &= \text{RNN}(\vh_{i-1}, \vy_{i-1}, \vc_i)
\end{align*}

Where the modified attention mechanism RecAtt follows:
\begin{align*}
  e_{ij} &= \vv^T\tanh\alpha(\vh_{i-1}, \vc_{i-1}, \vs_j) \\
  w_{ij} &= \frac{\text{exp}(e_{ij})}{\sum_{k}{\text{exp}(e_{ik})}} \\
  \vc_i &= \sum_{j}{w_{ij}\vs_j}
\end{align*}

The modified sum match function is:
\begin{align*}
  \alpha_{\text{sum}}(\va, \vb, \vc) = \vW^{\alpha}\va + \vU^{\alpha}\vb + \vV^{\alpha}\vc
\end{align*}

We note that RecAtt is purely content-based - the recurrent attention information is the context vector instead of the weights.  We show in our experiments that making attention unit itself recurrent is very important to improving end-to-end translation performance.

\subsection{RNNAtt}
\label{section:rnnatt}

\begin{figure}[t]
\centering
\includegraphics[width=0.9\columnwidth]{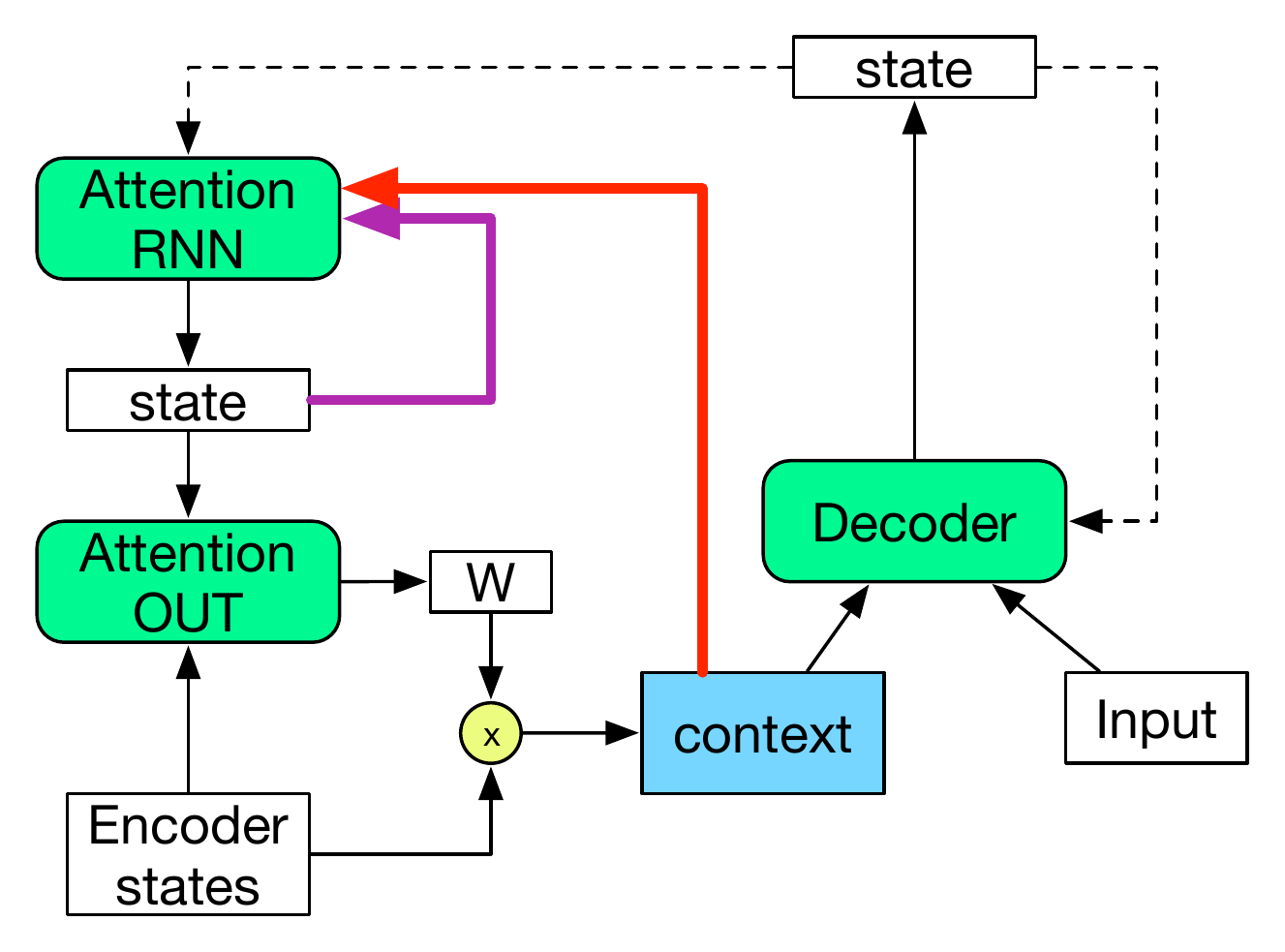}
\caption{\textsc{RNNAtt} Decoder.
    The purple thick line denotes the recurrent path of the hidden state of the attention unit.
    The red thick line denotes passing previous attention generated context to the attention unit
    for it to update its hidden state with respect to what information was extracted from the 
    source in the previous step.
}
\label{fig:rnnatt}
\end{figure}

RecAtt designed with the aim of adding a distortion model.
In RecAtt, only the previous attention-generated context is used in the recurrent path, so it only has a ``short-term memory".
To make it more flexible and have a longer memory, we propose \textsc{RNNAtt}, as shown in Fig.~\ref{fig:rnnatt}. The attention unit now keeps a hidden state and in effect becomes a complete RNN.
\begin{align*}
  \vc_i &= \text{ATT\_OUT}(\vq_{i-1}, \{\vs_j\}) \\
  \vq_i &= \text{ATT\_RNN}(\vq_{i-1}, \vh_{i-1}, \vc_i) \\
  \vh_i &= \text{RNN}(\vy_{i-1}, \vh_{i-1}, \vc_t)
\end{align*}

where ATT\_OUT is the original attention unit which applies the match function and softmax, ATT\_RNN denotes the hidden state $\vq_i$ computation of the attention unit.

\section{Conditioned Decoder}
\label{section:conditioned_decoder}

As analyzed in Sec.~\ref{section:problem_fertility}, attention mechanism might produce incorrect alignment and low-quality translation due to the lack of explicit distortion and fertility models. To address this issue, we propose conditioned decoder, \textsc{CondDec}, which uses a condition vector to represent what information has been extracted from the source sentence. This can be seen as an implicit fertility model, the condition vector can keep track of how many target words are translated from each source word. We use a structure similar to \cite{wen2015semantically} where a predefined condition is used to guide natural language generation. Different from that method, we use a trainable condition initialized with the last encoder hidden state. At each decoding step, the condition is updated with the decoder state and used to compute the next decoder state. The decoder GRU with attention and condition $\vsd_i$ is defined as adding an extra decay gate $vd_i$ to decoder:
\begin{align*}
  \vr_i &= \sigma(\vW^r\vx_{i} + \vU^r\vh_{i-1} + \vV^r\vc_i) \\
  \vh'_i &= \tanh(\vr_i \circ \vU\vh_{i-1} + \vW\vx_{i} + \vV\vc_i) \\
  \vz_i &= \sigma(\vW^z\vx_{i} + \vU^z\vh_{i-1} + \vV^z\vc_i) \\
  \vd_i &= \sigma(\vW^d\vx_{i} + \vU^d\vh_{i-1} + \vV^d\vc_i) \\
  \vsd_i &= \vd_i \circ \vsd_{i-1} \\
  \vh_i &= (1 - \vz_i) \circ \vh'_i + \vz_i \circ \vh_{i-1} + \tanh(\vV^h\vsd_i)
\end{align*}

Let $T$ be the length of the source sentence. We further penalize the condition by adding the following two costs to the categorical cross-entropy cost of the translation model:

\paragraph{Step-decay cost} We restrict the decay gate from extracting too much information from the condition. So we add a cost term:
\begin{align*}
\text{cost}_{decay} = \frac{1}{T}\sum_{j=1}^{T}||\vsd_j - \vsd_{j-1}||_2
\end{align*}

\paragraph{Left-over cost} We want the decoder to extract as much information as possible from the condition after reading the source sentence. So we add a cost term:
\begin{align*}
\text{cost}_{left} = ||\vsd_T||_2
\end{align*}

These two costs are added to the categorical cross-entropy cost of the translation model. At training time, the costs are used to enforce a fertility model and are ignored at test time.

\section{Related work}
There are variations of attention mechanism that have recurrent paths similar to that of \textsc{RecAtt}. In this section, we review these models and compare those decoder-attention structures.
\subsection{InputFeed \cite{luong2015effective}}
\label{section:inputfeed}

\begin{figure}[t]
\centering
\includegraphics[width=0.9\columnwidth]{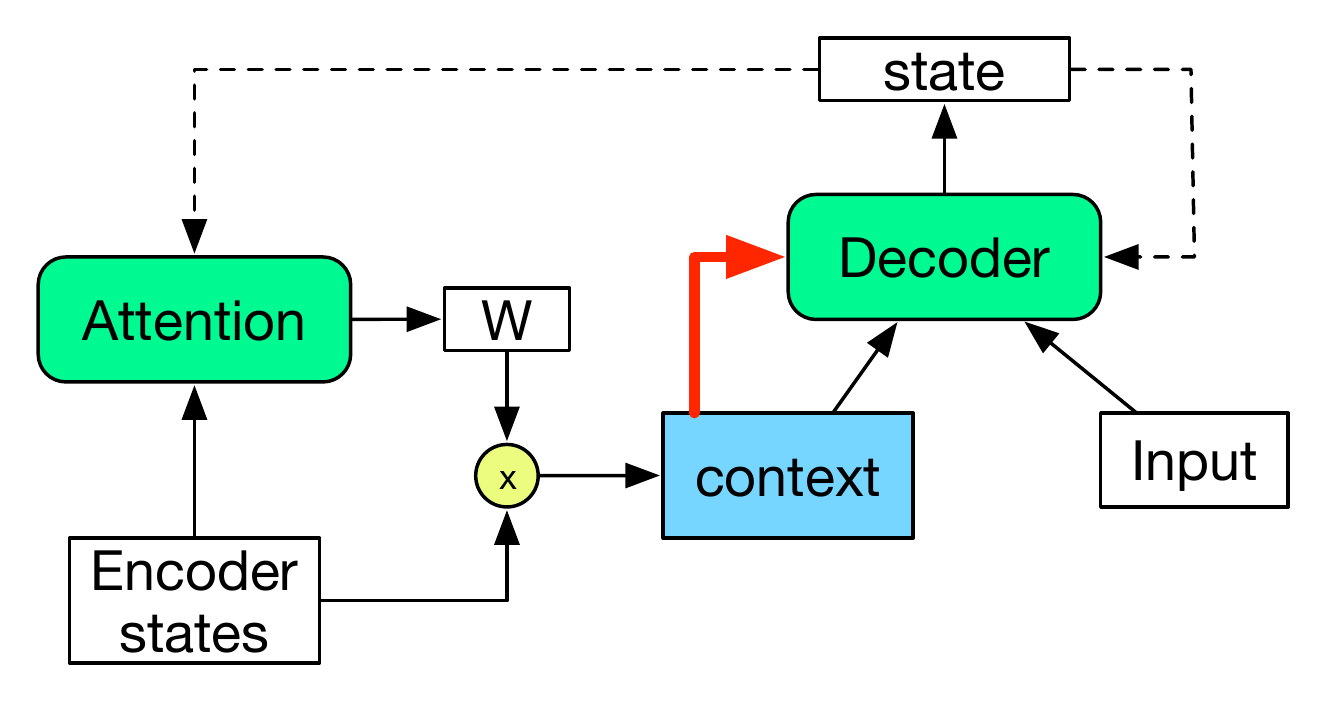}
\caption{\textsc{InputFeed} Decoder.
    The red thick line denotes the recurrent attention path which passes previous 
    attention generated context to the decoder.
}
\label{fig:inputfeed}
\end{figure}

In \cite{luong2015effective} the authors explored several variations of attention mechanism, including different match functions and local attention. We focus on the input-feeding method proposed in this paper because it is recurrent-like. \textsc{InputFeed} passes the previous attention output to the decoder together with current attention output, to further inform the decoder with previous alignment decisions.
\begin{align*}
  \vc_i &= \text{ATT}(\vh_{i-1}, \{\vs_i\}) \\
  \vh_t &= \text{RNN}(\vh_{i-1}, \vy_{i-1}, \vc_i, \vc_{i-1})
\end{align*}

This attention mechanism is purely content-based - the recurrent information is the context given by attention mechanism instead of weights. Note that the recurrent information is used outside the attention function, directly to the decoder, which makes it different from RecAtt, where the recurrent information is passed to the attention unit.

\subsection{HybridAtt1\cite{chorowski2014end}}
\label{section:hybridatt1}

\begin{figure}[t]
\centering
\includegraphics[width=0.9\columnwidth]{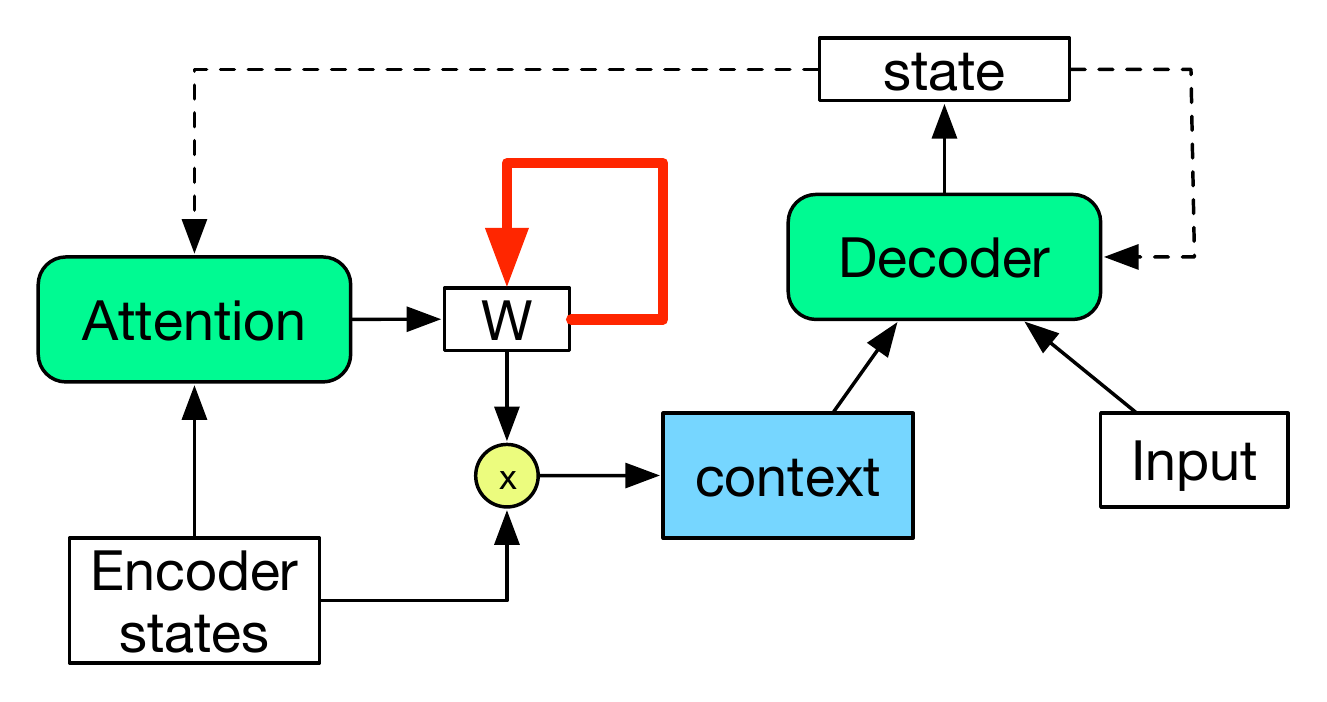}
\caption{\textsc{HybridAtt1} RNN Decoder.
    The red thick line denotes the recurrent attention path which uses previous 
    attention weights to adjust current attention weights.
}
\label{fig:hybridatt1}
\end{figure}

In \cite{chorowski2014end} the authors proposed an attention mechanism with a recurrent path. When computing the current set of weights on encoder states, the attention unit takes the previous weights and penalize the jump distance. It computes the average attention center, which is $m_{i-1}=\sum_{j}{j * w_{(i-1), j}}$. Then it adjusts the weight of each encoder state by its distance from that center. 
\begin{align*}
  m_{i-1} &= \sum_{j}{j \cdot w_{{i-1},j}} \\
  e_{ij} &= \vv^T \tanh\alpha(\vh_{i-1}, \vs_j) \\
  e'_{ij} &= \text{Logistic}(j - m_{i-1}) \cdot \text{exp}(e_{ij}) \\
  w_{ij} &= \frac{e'_{ij}}{\sum_{k}{e'_{ik}}} \\
  \vc_i &= \sum_{j}{w_{ij}\vs_j}
\end{align*}

This is a content-based attention with location-based recurrent attention, which is characterized by using an average attention center. Note that the recurrent information is used outside the attention unit, to adjusting the weights, which makes it different from RecAtt.

\subsection{HybridAtt2\cite{bahdanau2015end}}
\label{section:hybridatt2}

\begin{figure}[t]
\centering
\includegraphics[width=0.9\columnwidth]{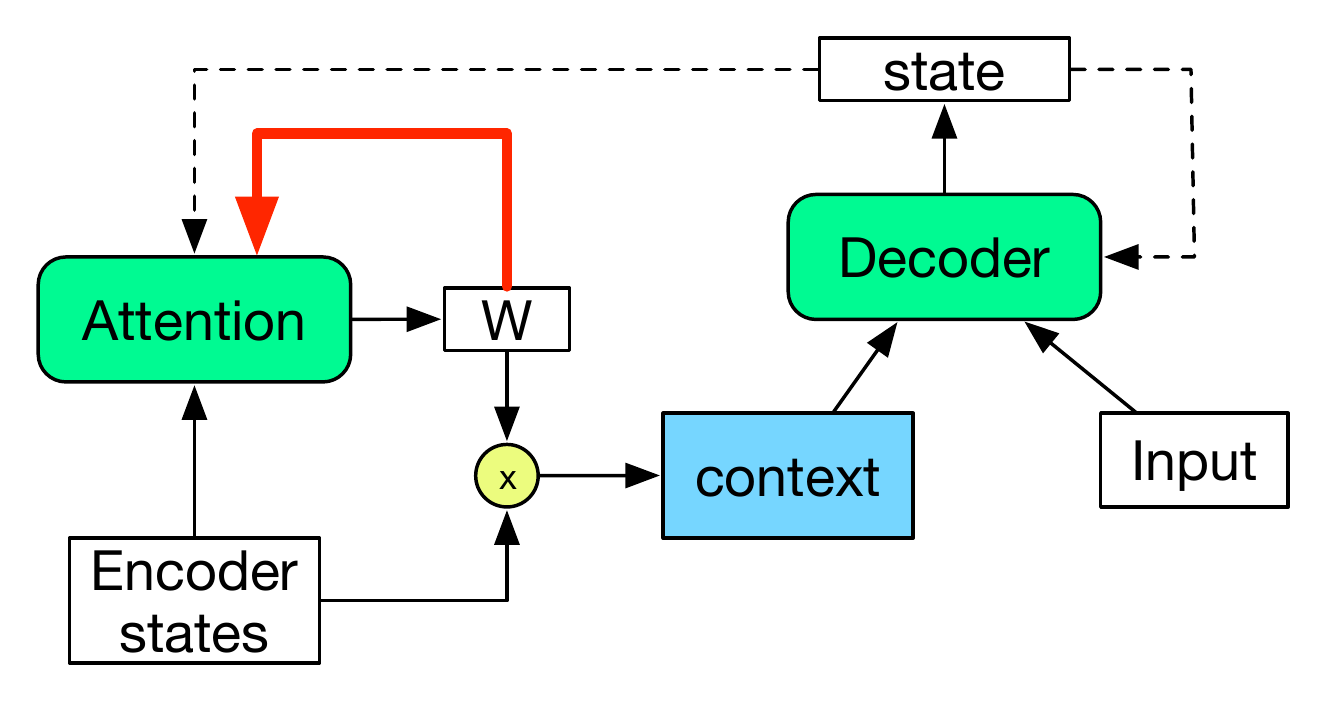}
\caption{\textsc{HybridAtt2} RNN Decoder.
    The red thick line denotes the recurrent attention path which passes previous 
    attention weights convoluted with $\vQ$ to the attention unit.
}
\label{fig:hybridatt2}
\end{figure}

In paper \cite{bahdanau2015end} the authors followed the previous one and improved \textsc{HybridAtt1} by integrating the recurrent location information into attention function. It first extracts feature vectors $\vg_i$ by doing convolution with previous weights $\vQ*w_{i-1}$, then uses these feature vectors to predict new weights.
\begin{align*}
  \vg_i &= \vQ * \vw_{i-1} \\
  e_{ij} &= \vv^T\tanh\alpha(\vh_{i-1}, s_j, \vg_{ij}) \\
  w_{ij} &= \frac{\text{exp}(e_{ij})}{\sum_{k}{\text{exp}(e_{ik})}} \\
  \vc_i &= \sum_{j}{w_{ij}\vs_j}
\end{align*}

where $*$ denotes convolution.

This is also a content-based attention mechanism with location-based recurrent attention. The difference between this method and \textsc{HybridAtt1} is that the recurrent information is integrated into the attention function.

We note that \textsc{HybridAtt1} and \textsc{HybridAtt2} were proposed for speech recognition task, which requires less re-ordering compared to translation. Although these models improved the performance of encoder-decoder on speech recognition, they not necessarily will on machine translation. We included these models because they have structures similar to \textsc{RecAtt}.

Other variations of attention mechanism with similar recurrent paths include \cite{mnih2014recurrent}, \cite{chen2015abc}. In these works, the authors used attention mechanism on image classification and visual question answering respectively. The variations of attention mechanism they used are location-based attention, which is more reasonable for image-related tasks. Due to these reasons we do not review or compare their methods in this work.

\section{Experimental Setup}
\label{section:exp_setup}

In this section we describe the data used in our experiments, our evaluation methods and our validation procedure. 

\begin{description}[leftmargin=0pt]

\item[Datasets]
For training, we use NIST Chinese-English training set excluding the Hong Kong Law and Hong Kong Hansard (0.5m sentence pairs after exclusion). For testing, we use Nist2005 dataset (1082 sentence pairs). For validation, we use Nist2003 dataset (913 sentence pairs). Validation set is only used for early-stopping and training process monitoring. 

Following \cite{bahdanau2014neural}, we use source and target dictionaries of size $30000$, covering 97.4\% and 98.9\% of the vocabularies respectively. Out-of-vocabulary words are replaced with a special token \unk.

\item[Post-processing]
We perform post-processing based on the alignment given by attention mechanism.  
For each translated target word, we choose the source word assigned with the highest attention weight as the aligned word.


\unk's in the translated sentence are replaced with the correct translation of the aligned source word. We make a simple word-level translation table from the alignment result given by GIZA++ \cite{och2003systematic} from the training set: for each source word, we choose the most frequently aligned target word.

\item[Evaluation] 

Performance is evaluated by BLEU score \cite{papineni2002bleu} over the test set. 

We compare 6 models, \textsc{RNNSearch} \cite{cho2014learning}, \textsc{HybridAtt2} \cite{bahdanau2015end}, \textsc{InputFeed} \cite{luong2015effective}, and three proposed models, \textsc{RecAtt}, \textsc{RNNAtt} and \textsc{CondDec}. We skip \textsc{HybridAtt1} because we have \textsc{HybridAtt2} as an improved version.

We benchmark the 6 NMT models with our implementation of hierarchical phrase-based SMT from \cite{chiang2007hierarchical}, with standard features, denoted as \textsc{SMT}.

\item[Validation]
Validation is done by calculating the BLEU score over the validation set without post-processing, using \texttt{MultiBleu.perl} script from \cite{bahdanau2014neural}.
For each model, we choose the parameters of the highest validation score.

\item[Model Training]
The encoder and decoder have $1000$ hidden units each. The dimension of source and target word embedding is $620$.
Following \cite{bahdanau2014neural}, we use dropout rate $0.5$.

We remove sentences of length over 50 words from the training set. 
We use batch size of 80 with 12 batches pre-fetched and sorted by the sentence length.

Each model is trained with AdaGrad \cite{duchi2011adaptive} on K40m GPU for approximately 4 days, finishing over 400000 updates, equivalent to 640 epochs.

When testing trained models, we use beam search \cite{graves2012sequence,boulanger2013audio,sutskever2014sequence} with beam size of 12.

\end{description}

\section{Results}
\label{section:results}

\subsection{Quantitative}
\label{section:quantitative}

\begin{table}
\centering
\begin{tabular}{l r r r}
           &	Before  &	After   & Improvement \\\hline
  \textsc{SMT}        &    /    & 32.25 & / \\
  \textsc{RNNSearch}  &	26.65	  &	31.02	& 4.37 \\\hline
  \textsc{HybridAtt2} &	24.60	  &	29.14	& 4.54 \\
  \textsc{InputFeed}  &	25.44	  &	29.02	& 3.58 \\\hline
  \textsc{RecAtt}     &	\textbf{28.10}	&	\textbf{33.14} & \textbf{5.04}	\\
  \textsc{RNNAtt}     &	25.04	  &	30.02	& 4.98 \\
  \textsc{CondDec}    & 27.48   & 32.21 & 4.73 \\\hline
\end{tabular}
\caption{BLEU scores w/o post-processing and the improvement from post-processing}
\label{table:overview_results}
\end{table}

BLEU scores on the test set are shown in (Table~\ref{table:overview_results}).

\textsc{RecAtt} performed best among NMT models, with and without post-processing. \textsc{RecAtt} achieved a $2.1$ BLEU score improvement over the original \textsc{RNNSearch}. 

Note that \textsc{RecAtt} also gained the most improvement from post-processing, $5.04$ BLEU points. In the post-processing, we use a naive translation table which is generated purely from the training data so the effect of post-processing depends largely on the quality of the alignment. Thus the gain from post-processing can be seen as a measurement of the quality of attention-generated alignment, and from this we see that \textsc{RecAtt} improved attention mechanism.

\textsc{CondDec} out-performed \textsc{RNNSearch} by 1 BLEU point, both with and without post-processing.

All three of our proposed models out-performed the phrase-based SMT baseline.

The combination of \textsc{CondDec} with \textsc{RecAtt} and \textsc{RNNAtt} is a work in progress.

\subsection{Qualitative}
\label{section:qualitative}

As mentioned in Sec.~\ref{section:problems}, the original attention-based encoder-decoder has some problems due to the lack of distortion and fertility models. In this section we will qualitatively evaluate how our models resolved these problems.

\begin{figure}[t]
\centering
\includegraphics[width=1.0\columnwidth]{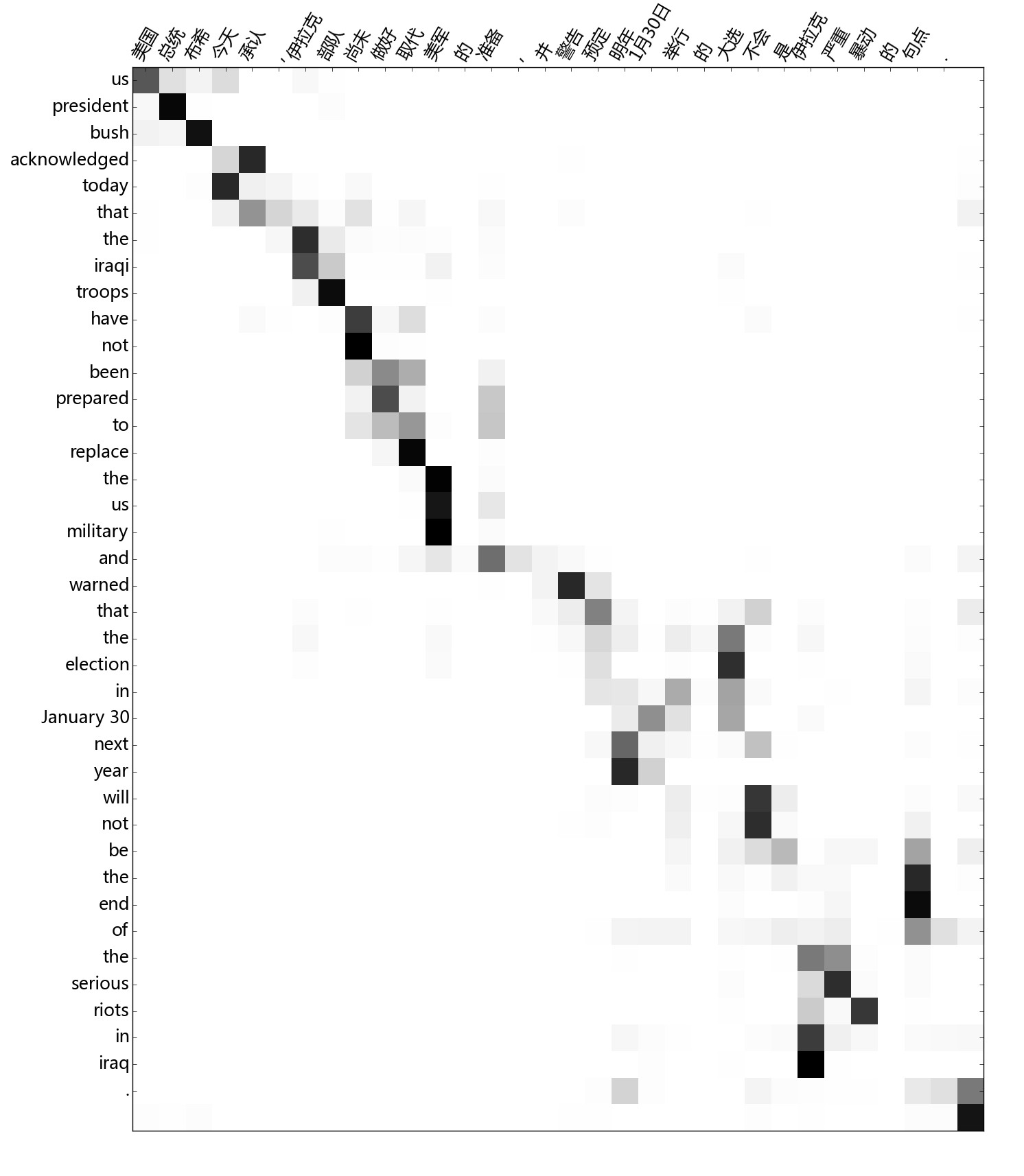}
\caption{Effect of implicit distortion model.}
\label{fig:expr_distortion_1}
\end{figure}

\paragraph{Distortion}

We show the alignment and translation by \textsc{RecAtt} in Fig.~\ref{fig:expr_distortion_1} on the same sentence of Fig.~\ref{fig:prob_distortion_1}. In the alignment by \textsc{RecAtt}, it can be seen that ``\emph{will not}" are correctly aligned to ``\begin{CJK}{UTF8}{gbsn}不会\end{CJK}" (means ``will not") and ``\emph{next year}" is correctly ordered to describe ``\emph{the election to be held}" instead of ``\emph{riot in iraq}". The transltion quality of the whole sentence is also higher.

\begin{figure}[t]
\centering
\includegraphics[width=1.0\columnwidth]{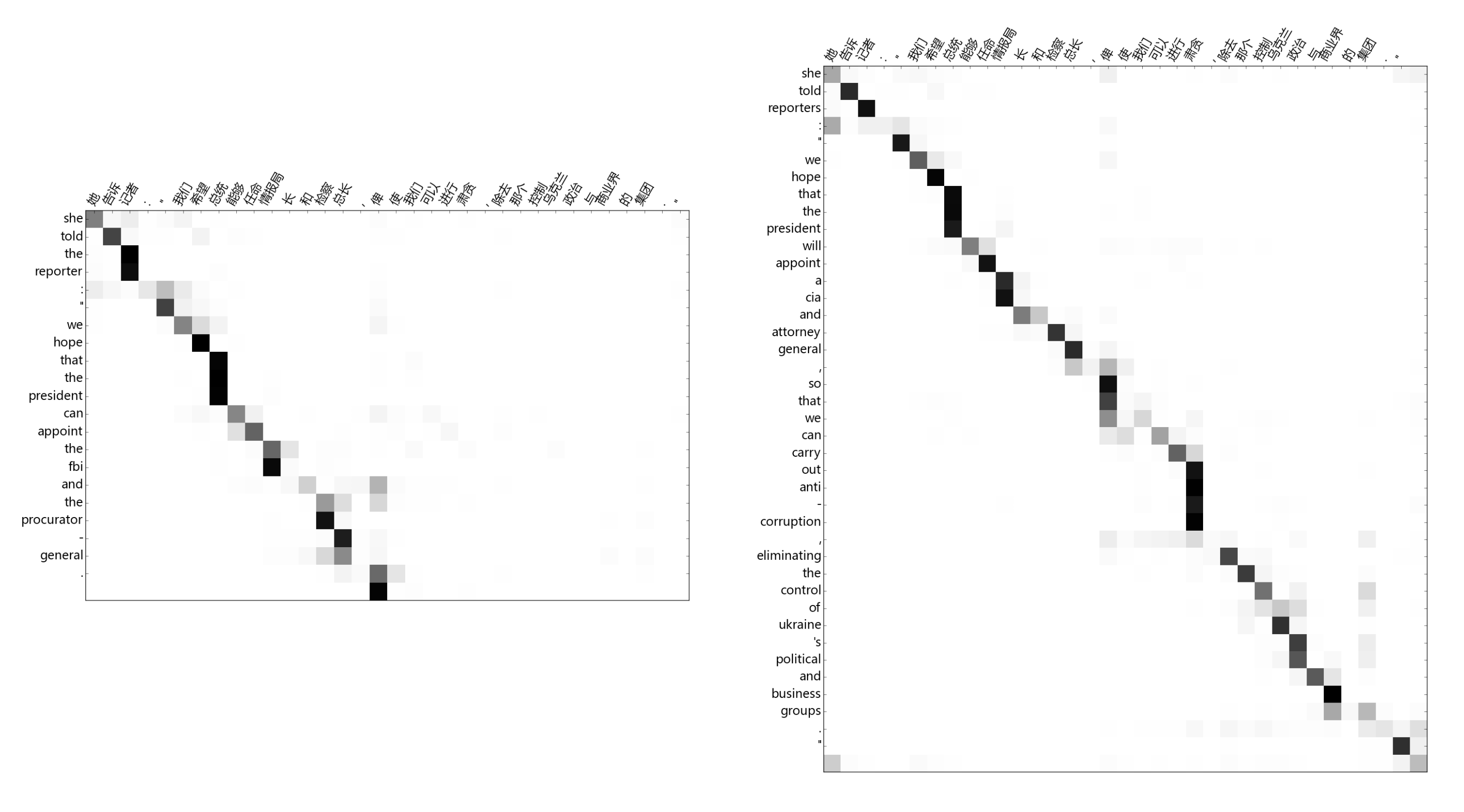}
\caption{Example of coverage problem. \\
Left: \textsc{RNNSearch}.
Right: \textsc{CondDec}.}
\label{fig:expr_coverage_1}
\end{figure}

\begin{figure}[t]
\centering
\includegraphics[width=1.0\columnwidth]{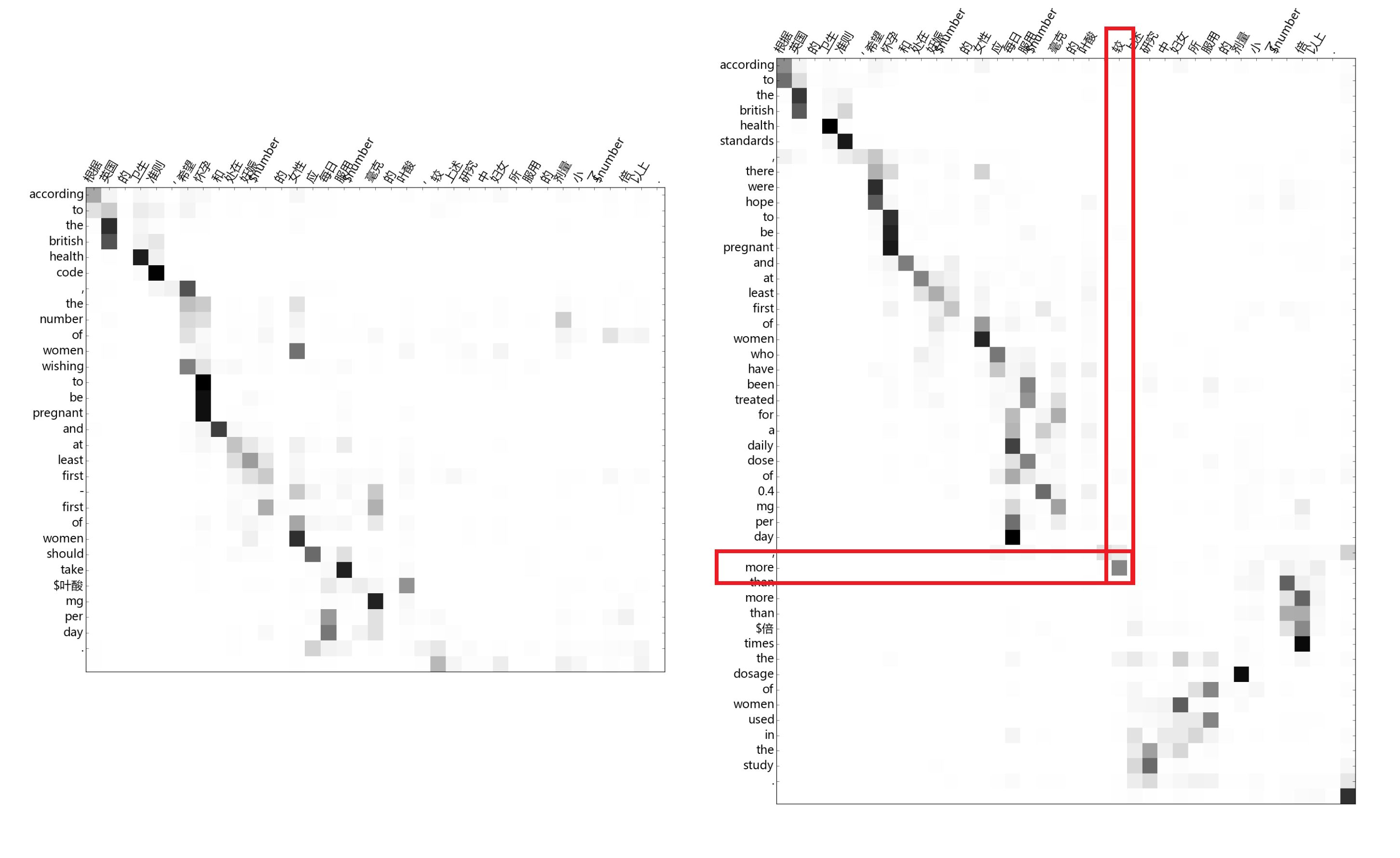}
\caption{Example of coverage problem. \\
Left: \textsc{RNNSearch}.
Right: \textsc{RecAtt}.}
\label{fig:expr_coverage_2}
\end{figure}

\paragraph{Fertility: Coverage} In Fig.~\ref{fig:expr_coverage_1} we show the alignments given by \textsc{RNNSearch} and \textsc{RecAtt}. From the alignment of \textsc{RNNSearch}, we can observe the problem of coverage where the later part of the source sentence is lost in the translation, while the alignment given by \textsc{RecAtt} does not have this problem and covered the whole source sentence. 

We observed that \textsc{RecAtt} can also resolve the coverage problem. This is because a correct alignment can be very helpful in preventing the incorrect generation of end-of-sentence symbol. In Fig.~\ref{fig:expr_coverage_2} we show an example. In the alignment by \textsc{RNNSearch}, when generating the word after ``," (last row), the attention is not very concentrated, leading to the generation of end-of-sentence symbol. While in the alignment by \textsc{RecAtt} when generating that word, the attention correctly focused on ``\begin{CJK}{UTF8}{gbsn}较\end{CJK}" (means ``more") with high confidence, leading to the correct generation of ``\emph{more}".

\begin{figure}[t]
\includegraphics[width=1.0\columnwidth]{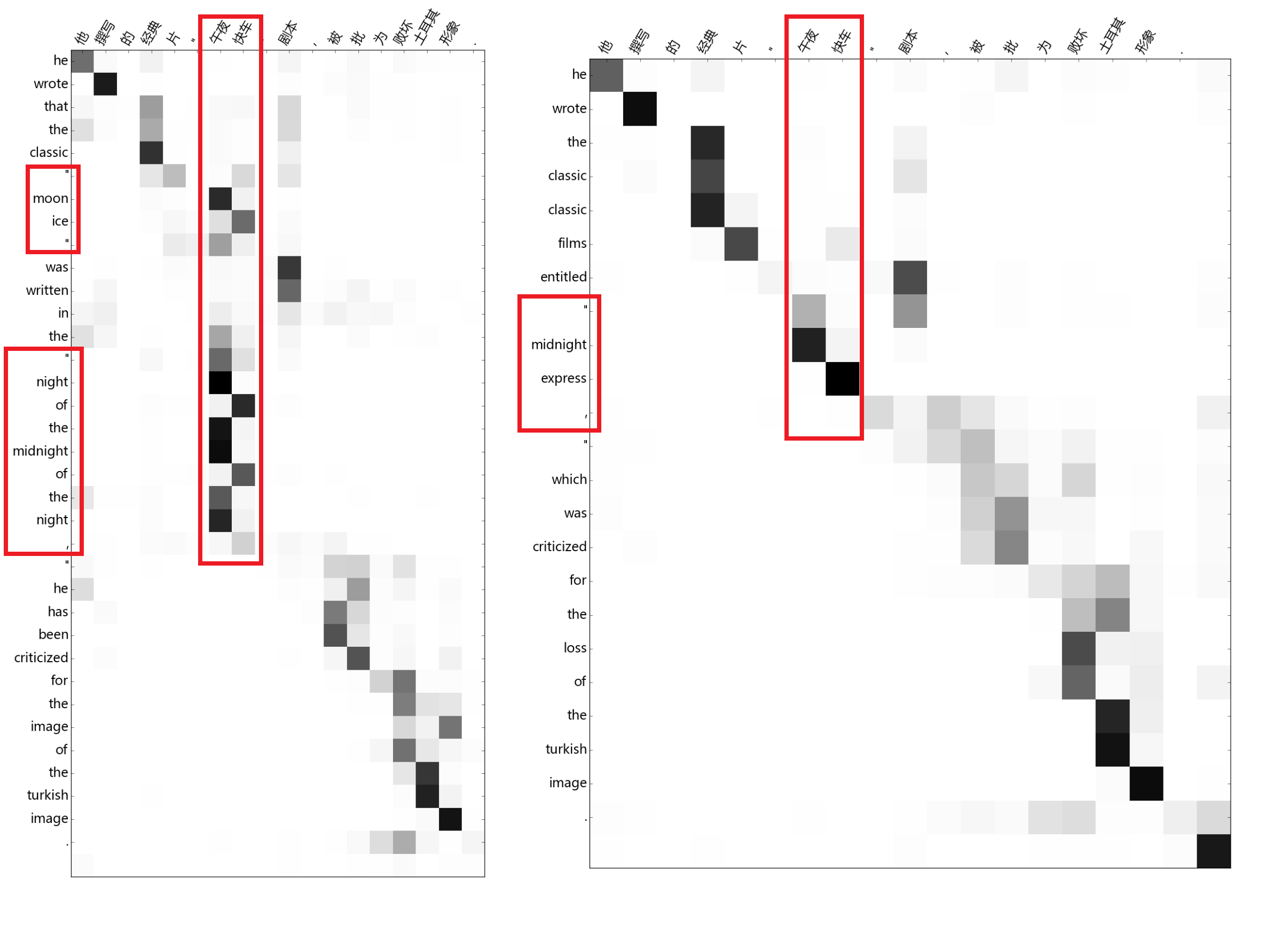}
\caption{Example of repetition problem. \\
Left: \textsc{RNNSearch}.
Right: \textsc{CondDec}.}
\label{fig:expr_repetition_1}
\end{figure}

\paragraph{Fertility: Repetition} In Fig.~\ref{fig:expr_repetition_1} we see that the problem of repetition occurred in the alignment by \textsc{RNNSearch}. ``\begin{CJK}{UTF8}{gbsn}东方 快车\end{CJK}" (means ``midnight express") is repeatedly focused on and translated into ``\emph{moon ice}" and ``\emph{night of the midnight of the night}". \textsc{CondDec} produces both the correct alignment and the correct translation ``\emph{midnight express}".

\begin{figure}[t]
\centering
\includegraphics[width=1.0\columnwidth]{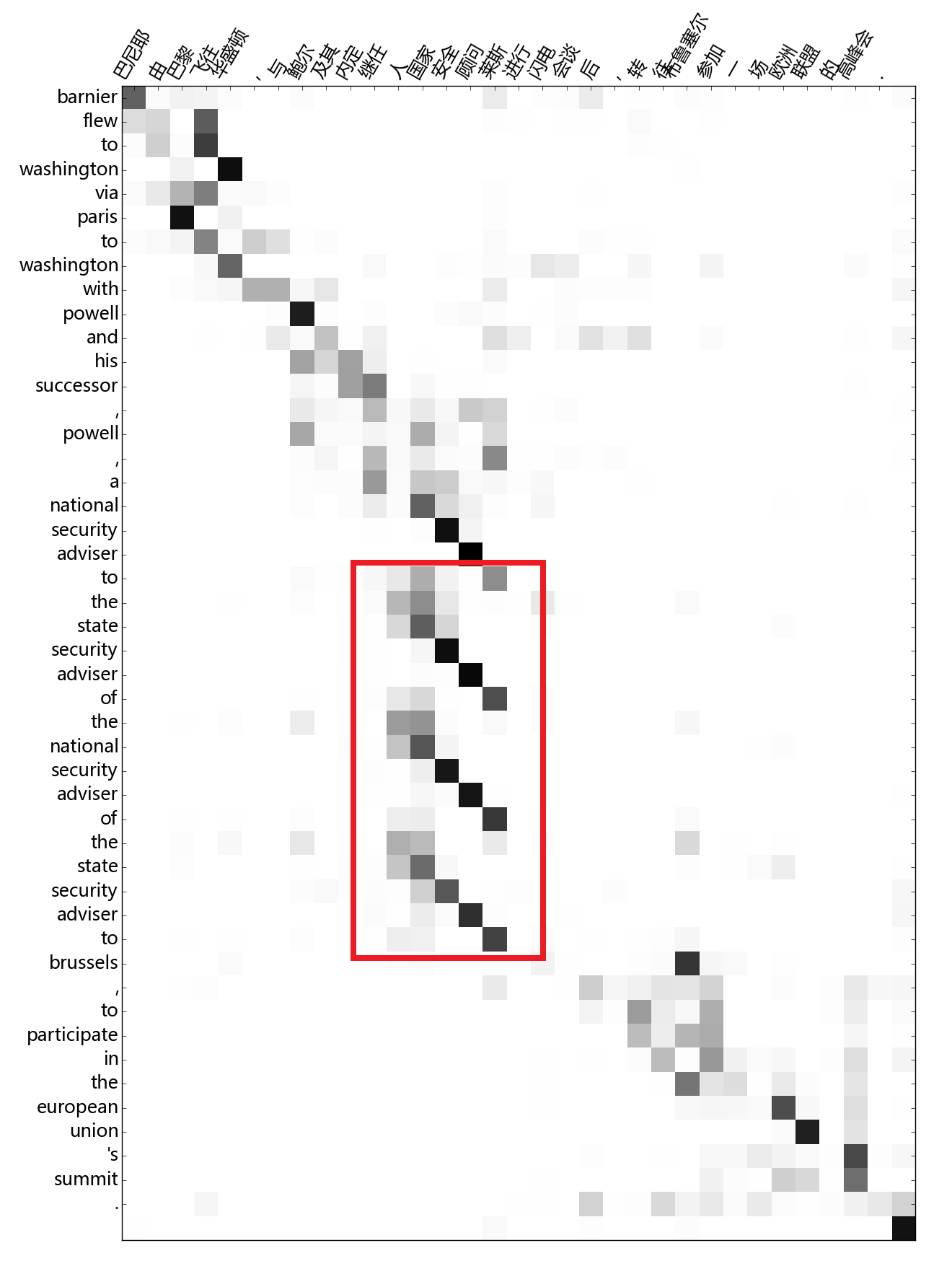}
\caption{Long repetition.}
\label{fig:expr_long_repetition_1}
\end{figure}

\begin{figure}[t]
\includegraphics[width=1.0\columnwidth]{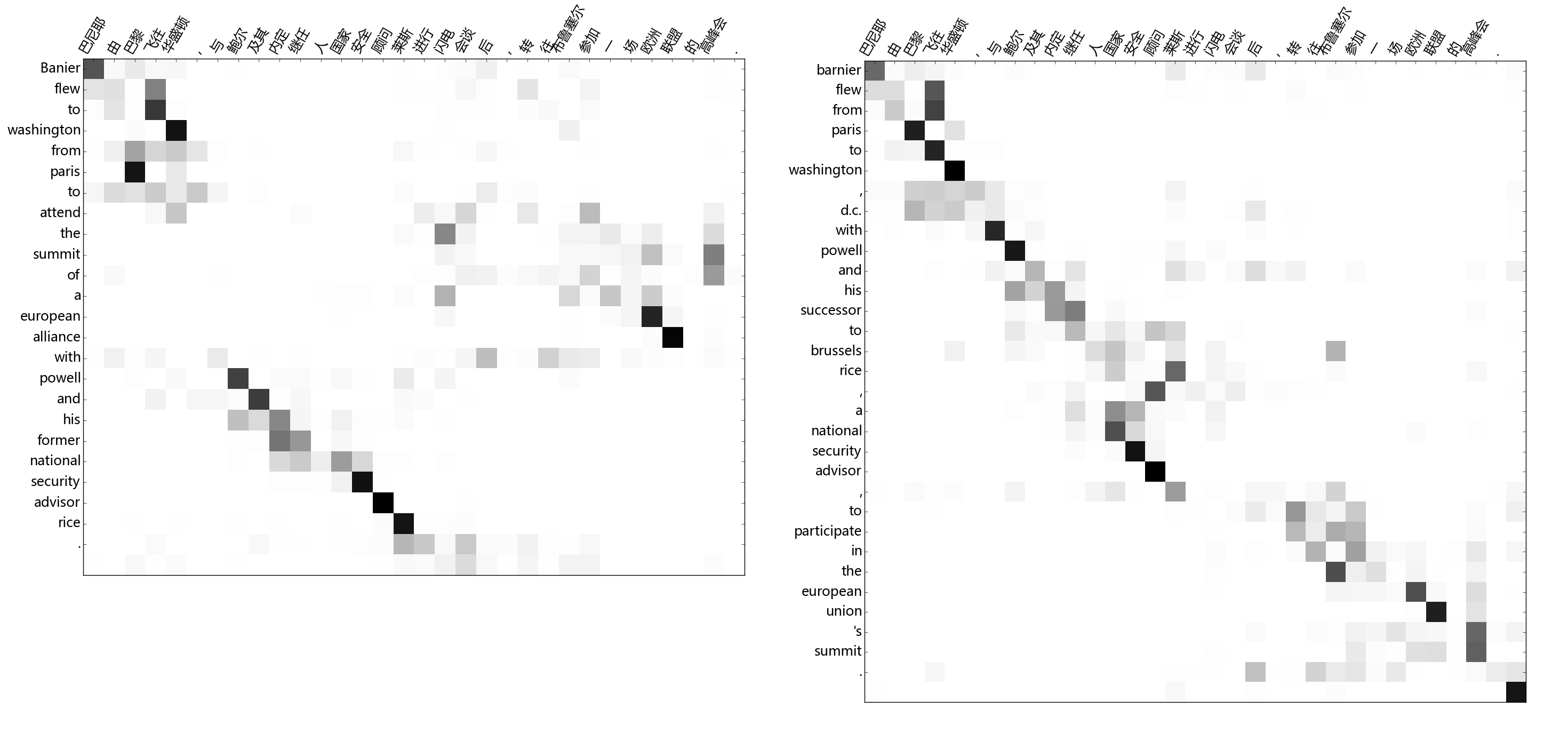}
\caption{Example of long repetition. \\
Left: \textsc{RNNSearch}.
Right: \textsc{RNNAtt}.}
\label{fig:expr_long_repetition_2}
\end{figure}

\paragraph{Long Repetition} 
We observed that \textsc{RecAtt} can also resolve the repetition problem. Because the previous attention-generated context was passed to the attention unit, the attention can decide not to focus on the same position as last time. But since it only has a short-term memory, in some cases the alignment by \textsc{RecAtt} has long repetitions as shown in Fig.~\ref{fig:expr_long_repetition_1}. 

Although \textsc{RNNAtt} did not perform as well as \textsc{RecAtt} in terms of BLEU score, we observe that it can resolve the long repetition problem that is hard for \textsc{RecAtt} to handle. In Fig.~\ref{fig:expr_long_repetition_2} we show the alignments by \textsc{RNNSearch} and \textsc{RNNAtt}. First we can see that \textsc{RNNSearch} did not handle this sentence well, with incorrect alignment and low-quality translation, which shows that this sentence is hard to translate. However the alignment by \textsc{RNNAtt} is more accurate and the translation quality is higher than both \textsc{RNNSearch} and \textsc{RecAtt}, with no long repetition problem.

One possible reason for the low BLEU score of \textsc{RNNAtt} is that the path from translation cost to the attention recurrent unit is too long and the model is hard to train. Improving end-to-end performance and exploring alternative structures for \textsc{RNNAtt} is a work in progress.



\section{Conclusions}

In this paper we noted some problems occurred in neural machine translation due to the lack of distortion and fertility model.

To resolve these problems, we proposed to add implicit distortion and fertility models to attention-based encoder-decoder. We proposed recurrent attention mechanism, \textsc{RecAtt} and \textsc{RNNAtt} for distortion model, and \textsc{CondDec} for fertility model. We compared our models with other related variations.

We evaluated our methods both quantitatively and qualitatively. In Chinese-English translation, \textsc{RecAtt} gained an improvement of 2 BLEU points over the original attention mechanism and out-performed all related models. \textsc{CondDec} also out-performed the original attention mechanism by 1 BLEU point.
By analyzing the alignment matrix produced by attention mechanism, we demonstrated that our proposed methods help resolve the observed problems.

We are working on the combination of \textsc{CondDec} with \textsc{RecAtt} and \textsc{RNNAtt}. We will also explore alternative structures of recurrent attention mechanism and try to improve the end-to-end performance of \textsc{RNNAtt}.


\bibliographystyle{acl2015}
\bibliography{master}

\begin{thebibliography}{}

\bibitem[\protect\citename{Bahdanau \bgroup et al.\egroup
  }2014]{bahdanau2014neural}
Dzmitry Bahdanau, Kyunghyun Cho, and Yoshua Bengio.
\newblock 2014.
\newblock Neural machine translation by jointly learning to align and
  translate.
\newblock {\em arXiv preprint arXiv:1409.0473}.

\bibitem[\protect\citename{Bahdanau \bgroup et al.\egroup
  }2015]{bahdanau2015end}
Dzmitry Bahdanau, Jan Chorowski, Dmitriy Serdyuk, Philemon Brakel, and Yoshua
  Bengio.
\newblock 2015.
\newblock End-to-end attention-based large vocabulary speech recognition.
\newblock {\em arXiv preprint arXiv:1508.04395}.

\bibitem[\protect\citename{Boulanger-Lewandowski \bgroup et al.\egroup
  }2013]{boulanger2013audio}
Nicolas Boulanger-Lewandowski, Yoshua Bengio, and Pascal Vincent.
\newblock 2013.
\newblock Audio chord recognition with recurrent neural networks.
\newblock In {\em ISMIR}, pages 335--340.

\bibitem[\protect\citename{Chen \bgroup et al.\egroup }2015]{chen2015abc}
Kan Chen, Jiang Wang, Liang-Chieh Chen, Haoyuan Gao, Wei Xu, and Ram Nevatia.
\newblock 2015.
\newblock Abc-cnn: An attention based convolutional neural network for visual
  question answering.
\newblock {\em arXiv preprint arXiv:1511.05960}.

\bibitem[\protect\citename{Chiang}2007]{chiang2007hierarchical}
David Chiang.
\newblock 2007.
\newblock Hierarchical phrase-based translation.
\newblock {\em computational linguistics}, 33(2):201--228.

\bibitem[\protect\citename{Cho \bgroup et al.\egroup }2014]{cho2014learning}
Kyunghyun Cho, Bart Van~Merri{\"e}nboer, Caglar Gulcehre, Dzmitry Bahdanau,
  Fethi Bougares, Holger Schwenk, and Yoshua Bengio.
\newblock 2014.
\newblock Learning phrase representations using rnn encoder-decoder for
  statistical machine translation.
\newblock {\em arXiv preprint arXiv:1406.1078}.

\bibitem[\protect\citename{Chorowski \bgroup et al.\egroup
  }2014]{chorowski2014end}
Jan Chorowski, Dzmitry Bahdanau, Kyunghyun Cho, and Yoshua Bengio.
\newblock 2014.
\newblock End-to-end continuous speech recognition using attention-based
  recurrent nn: First results.
\newblock {\em arXiv preprint arXiv:1412.1602}.

\bibitem[\protect\citename{Duchi \bgroup et al.\egroup
  }2011]{duchi2011adaptive}
John Duchi, Elad Hazan, and Yoram Singer.
\newblock 2011.
\newblock Adaptive subgradient methods for online learning and stochastic
  optimization.
\newblock {\em The Journal of Machine Learning Research}, 12:2121--2159.

\bibitem[\protect\citename{Goodfellow \bgroup et al.\egroup
  }2013]{goodfellow2013maxout}
Ian~J Goodfellow, David Warde-Farley, Mehdi Mirza, Aaron Courville, and Yoshua
  Bengio.
\newblock 2013.
\newblock Maxout networks.
\newblock {\em arXiv preprint arXiv:1302.4389}.

\bibitem[\protect\citename{Graves}2012]{graves2012sequence}
Alex Graves.
\newblock 2012.
\newblock Sequence transduction with recurrent neural networks.
\newblock {\em arXiv preprint arXiv:1211.3711}.

\bibitem[\protect\citename{Hermann \bgroup et al.\egroup
  }2015]{hermann2015teaching}
Karl~Moritz Hermann, Tomas Kocisky, Edward Grefenstette, Lasse Espeholt, Will
  Kay, Mustafa Suleyman, and Phil Blunsom.
\newblock 2015.
\newblock Teaching machines to read and comprehend.
\newblock In {\em Advances in Neural Information Processing Systems}, pages
  1684--1692.

\bibitem[\protect\citename{Hochreiter and Schmidhuber}1997]{hochreiter1997long}
Sepp Hochreiter and J{\"u}rgen Schmidhuber.
\newblock 1997.
\newblock Long short-term memory.
\newblock {\em Neural computation}, 9(8):1735--1780.

\bibitem[\protect\citename{Luong \bgroup et al.\egroup
  }2015]{luong2015effective}
Minh-Thang Luong, Hieu Pham, and Christopher~D Manning.
\newblock 2015.
\newblock Effective approaches to attention-based neural machine translation.
\newblock {\em arXiv preprint arXiv:1508.04025}.

\bibitem[\protect\citename{Mnih \bgroup et al.\egroup }2014]{mnih2014recurrent}
Volodymyr Mnih, Nicolas Heess, Alex Graves, et~al.
\newblock 2014.
\newblock Recurrent models of visual attention.
\newblock In {\em Advances in Neural Information Processing Systems}, pages
  2204--2212.

\bibitem[\protect\citename{Och and Ney}2003]{och2003systematic}
Franz~Josef Och and Hermann Ney.
\newblock 2003.
\newblock A systematic comparison of various statistical alignment models.
\newblock {\em Computational linguistics}, 29(1):19--51.

\bibitem[\protect\citename{Papineni \bgroup et al.\egroup
  }2002]{papineni2002bleu}
Kishore Papineni, Salim Roukos, Todd Ward, and Wei-Jing Zhu.
\newblock 2002.
\newblock Bleu: a method for automatic evaluation of machine translation.
\newblock In {\em Proceedings of the 40th annual meeting on association for
  computational linguistics}, pages 311--318. Association for Computational
  Linguistics.

\bibitem[\protect\citename{Pascanu \bgroup et al.\egroup
  }2012]{pascanu2012difficulty}
Razvan Pascanu, Tomas Mikolov, and Yoshua Bengio.
\newblock 2012.
\newblock On the difficulty of training recurrent neural networks.
\newblock {\em arXiv preprint arXiv:1211.5063}.

\bibitem[\protect\citename{Shih \bgroup et al.\egroup }2015]{shih2015look}
Kevin~J Shih, Saurabh Singh, and Derek Hoiem.
\newblock 2015.
\newblock Where to look: Focus regions for visual question answering.
\newblock {\em arXiv preprint arXiv:1511.07394}.

\bibitem[\protect\citename{Sutskever \bgroup et al.\egroup
  }2014]{sutskever2014sequence}
Ilya Sutskever, Oriol Vinyals, and Quoc~VV Le.
\newblock 2014.
\newblock Sequence to sequence learning with neural networks.
\newblock In {\em Advances in neural information processing systems}, pages
  3104--3112.

\bibitem[\protect\citename{Vinyals \bgroup et al.\egroup
  }2014]{vinyals2014show}
Oriol Vinyals, Alexander Toshev, Samy Bengio, and Dumitru Erhan.
\newblock 2014.
\newblock Show and tell: A neural image caption generator.
\newblock {\em arXiv preprint arXiv:1411.4555}.

\bibitem[\protect\citename{Wen \bgroup et al.\egroup
  }2015]{wen2015semantically}
Tsung-Hsien Wen, Milica Gasic, Nikola Mrksic, Pei-Hao Su, David Vandyke, and
  Steve Young.
\newblock 2015.
\newblock Semantically conditioned lstm-based natural language generation for
  spoken dialogue systems.
\newblock {\em arXiv preprint arXiv:1508.01745}.

\bibitem[\protect\citename{Xu and Saenko}2015]{xu2015ask}
Huijuan Xu and Kate Saenko.
\newblock 2015.
\newblock Ask, attend and answer: Exploring question-guided spatial attention
  for visual question answering.
\newblock {\em arXiv preprint arXiv:1511.05234}.

\bibitem[\protect\citename{Xu \bgroup et al.\egroup }2015]{xu2015show}
Kelvin Xu, Jimmy Ba, Ryan Kiros, Aaron Courville, Ruslan Salakhutdinov, Richard
  Zemel, and Yoshua Bengio.
\newblock 2015.
\newblock Show, attend and tell: Neural image caption generation with visual
  attention.
\newblock {\em arXiv preprint arXiv:1502.03044}.

\end{thebibliography}

\end{document}